\documentclass[10pt,twocolumn]{article} 
\usepackage{simpleConference}
\usepackage{cite}
\usepackage{amsmath,amssymb,amsfonts}
\usepackage{algorithmic}
\usepackage{graphicx}
\usepackage{textcomp}

\usepackage[utf8]{inputenc} 
\usepackage[T1]{fontenc}    
\usepackage{hyperref}       
\usepackage{url}            
\usepackage{booktabs}       
\usepackage{amsfonts}       
\usepackage{nicefrac}       
\usepackage{microtype}      
\usepackage{lipsum}
\usepackage{framed,multirow}
\usepackage{amssymb}
\usepackage{latexsym}
\usepackage{makecell}
\usepackage{url}
\usepackage{xcolor}
\definecolor{newcolor}{rgb}{.8,.349,.1}
\usepackage{ragged2e}
\usepackage{color,soul}
\usepackage{hyperref}
\usepackage{graphicx}
\usepackage{caption}
\usepackage{subcaption}
\usepackage{float}

\usepackage{array}
\newcolumntype{P}[1]{>{\centering\arraybackslash}p{#1}}

\begin{document}

\title{A Convolutional Neural Network Approach to the Classification of Engineering Models}

\author{\textbf{Bharadwaj Manda, Pranjal Bhaskare, Ramanathan Muthuganapathy} \\
\small{Indian Institute of Technology Madras}}

\maketitle
\thispagestyle{empty}

\begin{abstract}
This paper presents a deep learning approach for the classification of Engineering (CAD) models using Convolutional Neural Networks (CNNs). Owing to the availability of large annotated datasets and also enough computational power in the form of GPUs, many deep learning-based solutions for object classification have been proposed of late, especially in the domain of images and graphical models. Nevertheless, very few solutions have been proposed for the task of functional classification of CAD models. Hence, for this research, CAD models have been collected from Engineering Shape Benchmark (ESB), National Design Repository (NDR) and augmented with newer models created using a modelling software to form a dataset - `CADNET'. It is proposed to use a residual network architecture for CADNET, inspired by the popular ResNet. A weighted Light Field Descriptor (LFD) scheme is chosen as the method of feature extraction, and the generated images are fed as inputs to the CNN. The problem of class imbalance in the dataset is addressed using a class weights approach. Experiments have been conducted with other signatures such as geodesic distance etc. using deep networks as well as other network architectures on the CADNET.  The LFD-based CNN approach using the proposed network architecture, along with gradient boosting yielded the best classification accuracy on CADNET.
\end{abstract}


\section{Introduction}
\label{sec:introduction}
Classification of Engineering (CAD) models is very important for a task such as design reuse. It has been observed that designers spend a considerable amount of time in search for the right information as well as use a large percentage of existing design for a new product development \cite{Gunn82}. Gunn \mbox{\cite{Gunn82}} has observed that about 40\% of the new designs could be built from an existing design and 40\% from modifying an existing design. Ullman \mbox{\cite{Ullman10}} has indicated that a large percentage (75\% or sometimes, more than that) of design reuses existing knowledge for the new product development. Classification is also an important task for retrieval of CAD models, which in turn employed in design reuse \mbox{\cite{BAI20101069}}. Another area of interest is in the CAD assembly model retrieval \mbox{\cite{CAD_assem}}, where, apart from using topology and connection informations, classification plays a key role. The interdependence between product life cycle management (PLM), material requirements planning (MRP) and CAD systems also calls for classification and search of 3D Engineering models \mbox{\cite{ShapeSearch}}. Considering the applications and the fact that we are in the digital age with many information archived digitally, the problem of automatic classification of CAD models becomes a predominant one.

\begin{figure}[t]
    \centering
    \begin{subfigure}[b]{0.2\textwidth}
        \centering
        \includegraphics[height=3.2cm,width=4cm]{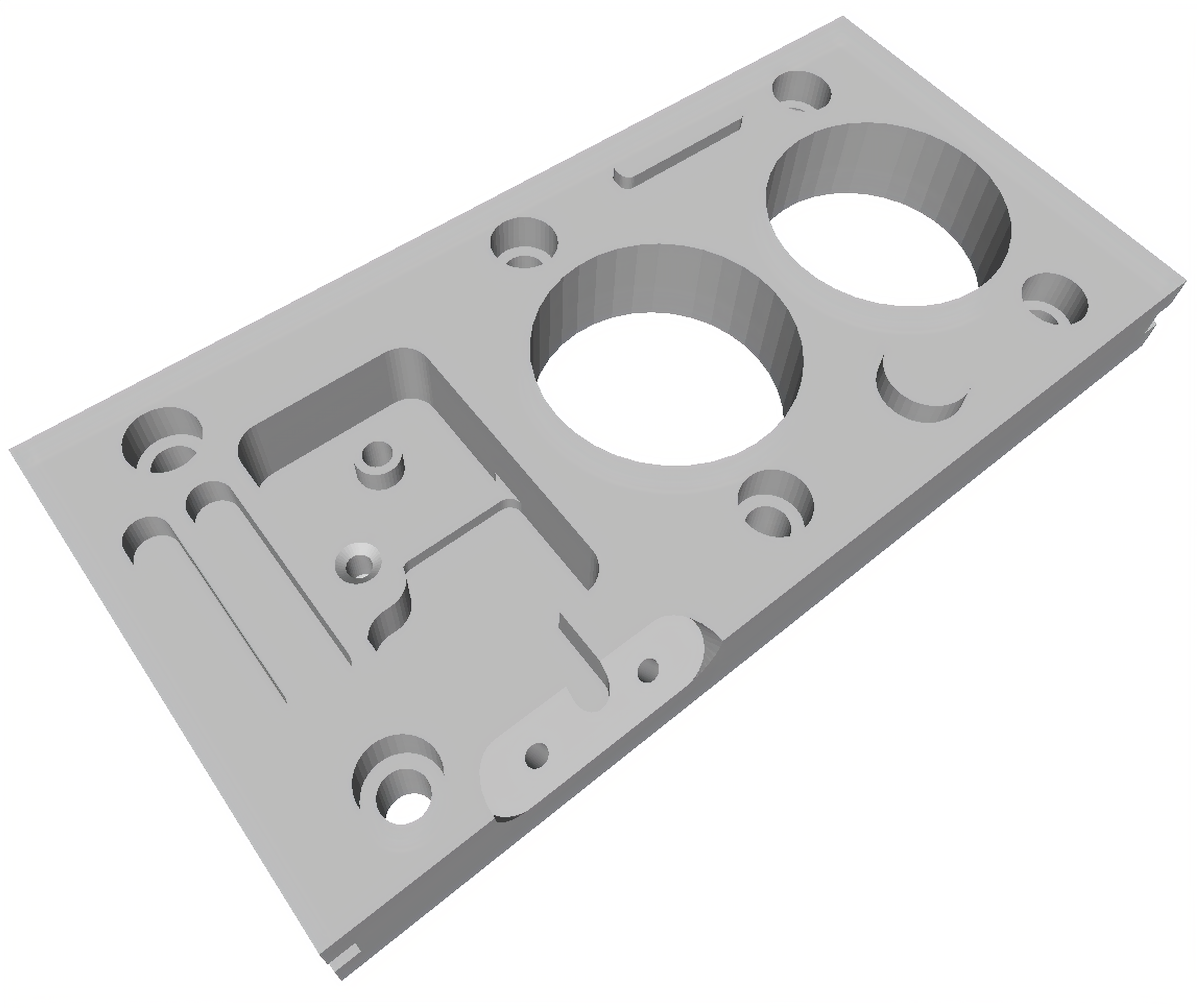}
        \caption{A 3D CAD Model}
        \label{fig1a}
    \end{subfigure}\qquad
    \begin{subfigure}[b]{0.2\textwidth}
        \centering
        \includegraphics[scale=0.15]{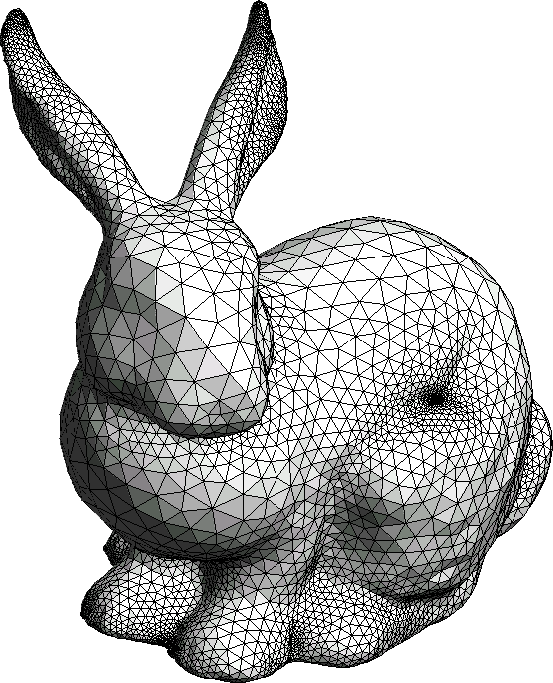}
        \caption{A 3D Graphical Model}
        \label{fig1b}
    \end{subfigure}
    \caption{Distinction between a CAD Model and a Graphical Model}
\end{figure}

An Engineering/CAD model (see Fig. \ref{fig1a}) has features such as holes (genus $>$ 0), blind holes (genus $=$ 0) and fillets which are usually absent in a graphical model (see Fig. \ref{fig1b}). Also, sharp edges are usually found in a CAD Model as opposed to a graphical model which more or less has smooth curvature throughout.

Traditionally, in the field of Engineering models, boundary representation (B-Rep) is the native format to store the data. To protect the proprietary design during data transfer, other formats such as mesh / tessellated representations are used. Also, with newer technologies such as additive manufacturing / 3D printing, mesh model representation in the field of Engineering / CAD is gaining popularity. 

A mesh for a CAD model (see Fig. \ref{fig1a}) is substantially different from that of a graphical model (see Fig. \ref{fig1b}) in the following ways \cite{ESB}:

\begin{itemize}
    \item CAD Model tessellations are typically sparse while that of graphical models are usually dense.
    \item CAD Models have a lesser number of triangles in general as compared to a graphical model.
\end{itemize}

In general, the problem of classification of shapes/models has been an active area of research in several fields viz. images, graphical models, CAD models etc. for more than two decades. The problem gained prominence with the start of digital archiving. For image data,  MPEG dataset \cite{mpeg7} was perhaps one of the first ones in the domain of computer vision. In the area of graphical models, Princeton shape benchmark (PSB) was one of the earlier ones \cite{psb}. Jayanti et al. introduced Engineering shape benchmark (ESB) for CAD models \cite{ESB}. The National Design Repository (NDR) \cite{NDR} also contains a few hundred CAD models. 

The advent of machine learning techniques and in particular, the advances made in deep learning, accelerated the research on the problem of classification. However, deep learning techniques call for a large number of labelled data with appropriate class information. Hence, labelled datasets for images with class information have grown much larger \mbox{\cite{Russakovsky2015}}, and so for graphical 3D models \mbox{\cite{ShapeNet}}. In recent times, datasets for Engineering/CAD models have also grown through acquiring from different resources \cite{ABC}. Though the dataset \cite{ABC} contains a large number of models, the aim seems to be more on populating the data rather than providing classification. In the field of Engineering, it is imperative to classify data functionally. For example, a pipe and a bolt may look like a cylinder but have different functionality. This classification task also requires the people involved to have rich domain knowledge and experience. As CAD models are a derivate of the Engineering design process, many of the design data are also proprietary in nature and hence may not be put in the public domain \cite{Qin2014}.  Also, there exists only very few works on CAD model classification using deep learning. Qin et al. \cite{Qin2014} use only deep networks (not CNN) and  the work presented in \cite{ABC} uses basic geometric properties such as normals and curvatures but does not take functional classification into account. 

Our motivation for addressing the problem of classification of Engineering / CAD models comes from the following:
\begin{enumerate}
\item Most of the CAD datasets (such as \mbox{\cite{ESB}} or \mbox{\cite{NDR}}) have only a few hundred models. 
\item Datasets having larger number of CAD models are either proprietary  (not publicly available) \mbox{\cite{Qin2014}} or lack classification information \mbox{\cite{ABC}}.
\item The recent advances in deep learning such as CNN have not been made use of, to the best of our knowledge. 
\end{enumerate}

In this work, in order to classify the CAD data functionally, we start by using the publicly available datasets of CAD Models, ESB and NDR, which also have well-annotated functional classification. Unfortunately, they have only very few models, in the order of hundreds. We then resort to creating CAD models and functionally classify them by adding to the appropriate class. A dataset termed `CADNET' has been then prepared.  A Convolutional Neural Network (CNN) approach (only deep network was used in \cite{Qin2014}) for the classification of CAD models is then proposed. It is also crucial to come up with a network architecture for the intended application. The key contributions of the paper are as follows: 

\begin{enumerate}
	\item A dataset named `CADNET', which is suitable for deep learning-based approaches.
    \item A CNN-based deep learning approach for the classification of CAD models using a residual network structure, inspired by ResNet \cite{ResNet}, with much lesser number of filters and thereby much reduction in the number of parameters.
    \item Used the idea of class-weights, in order to alleviate the problem of imbalanced classes in CADNET.
    \item Proposed the use of gradient-based boosting approaches to improve the efficiency of classification.
    \item The proposed network produces better classification accuracy with much lesser training time. 
\end{enumerate}  

The manuscipt is organised as follows: Section \ref{sec2} discusses the literature corresponding to 3D CAD models, in addtion to the literature on Images, 3D Graphical Models, and an overview of the existing datasets for CAD model classification. The typical pipeline to be employed for a deep learning based classification task is described in Section \ref{sec3}, and each step of the pipeline is elaborated in Sections \ref{sec4} to \ref{sec6}. Section \ref{sec7} provides the Implementation Details. The results, limitaitons and possible future work are elaborated in Section \ref{sec:res}, followed by a Conclusion (Section \ref{sec:concl}).

\section{Related Works}
\label{sec2}
Many works in recent times have focussed on 3D graphical models and images. However, we focus more on the approaches that have been proposed for the task of classifying 3D CAD models, which are very few. 

\renewcommand{\arraystretch}{1.2}
\begin{table*}[!h]
\resizebox{\textwidth}{!}{%
    \centering
    \begin{tabular}{|P{2cm}|P{2cm}|P{2cm}|P{2cm}|P{7cm}|}
        \hline
        \textbf{Reference} & \textbf{Total Num. of Models} & \textbf{Num. of classes} & \textbf{Avg. Num. of models per class} & \textbf{Comments} \\
        
        \hline
        Wu et al. \cite{Wu} & \centering 36 & \centering -- & \centering -- & (-) contains only one class \\
        
        \hline
        Ip et al. \cite{IpRegli}; \& \newline Ip et al. \cite{IpRegli2}& \centering 85 & \centering 12 & \centering 7 & (+) functional classification \newline (-) less number of models; not suitable for deep learning;  \\
        
        \hline
        Ip et al. \cite{IpRegli}; \& \newline Ip et al. \cite{IpRegli2}& \centering 56 & \centering 4 & \centering 14 & (-) too few models \newline (-) manufacturing classification; \\
        
        \hline
        Ip et al. \cite{IpRegli3}& \centering 100 & \centering -- & \centering -- & (-) single category; \newline (-) manufacturing classification; \\
        
        \hline
        Hou et al. \cite{Hou}& \centering 218 & \centering 6 & \centering 36 & (+) functional classification \newline (-) too few classes and less number of models; \\
        
        \hline
        Bespalov et al. \cite{NDR} (NDR - Functional) & \centering 70 & \centering 10 & \centering 7 & (+) functional classification\newline (-) too few models per class;\\
        
        \hline
        Jayanti et al. \cite{ESB} (ESB) & \centering 801 & \centering 42 & \centering 19 & (+) functional classification \newline (+) reasonably sized;\\
        
        \hline
        Qin et al. \cite{Qin2014}& \centering 7464 & \centering 28 & \centering 266 & (+) functional classification \newline (+) large dataset, suitable for deep learning \newline (-) dataset unavailable;\\
        
        \hline
        Koch et al. \cite{ABC}& \centering 1 million & \centering - & \centering - & (+) large dataset \& available \newline (-) lack of any category or class information makes it unsuitable for classification;\\
        
        \hline
        \multicolumn{5}{@{}r}{`+' -- Advantages; `-' -- Limitations}
    \end{tabular}
    }
    \caption{\label{tab1} Summary of existing datasets indicated the advantages and limitations of each of them. Most of the datasets are small as indicated by their respective numbers.}
\end{table*}
\renewcommand{\arraystretch}{1.00}

\subsection{3D CAD models}

Wu and Jen \cite{Wu} proposed a neural network approach to classify 3D prismatic parts. The idea used the hypothesis that a 3D part could be modelled by the contours of its three projected views. The views were then approximated by as many rectilinear polygons. Such a representation was used as an input vector to a back-propagation neural network. A total of 36 parts were classified in this way but the classification was hierarchical and not functional. The research presented in \cite{IpRegli} and \cite{IpRegli2} aimed at performing a classification of the CAD models based on their manufacturing process as well as their functionality. The input CAD mesh models were converted into a histogram representation using enhanced shape distribution, and the extracted representation was then fed into a $k$-Nearest Neighbour ($k$NN) classifier. Support Vector Machines (SVMs) were employed for classification approach using surface curvatures as a feature in \cite{IpRegli3}, although the classification was not based on model functionality. The idea of SVMs was also used by \cite{Hou}, where a hybrid of moment invariants, principal moments and geometric ratios were used as input feature vectors. In each of these studies, the classification accuracy was not very high. A comparison of some of the shape signatures has been provided in \cite{ESB} for classification of CAD models. 

\subsection{Images and 3D graphical models}

For images and 3D graphical models, there exists a plethora of literature in the last few years that employed advanced deep learning techniques such as Convolutional Neural Networks (CNNs). We mention only a few for the sake of completeness (it may be noted that neural networks have been employed in other fields such as control systems, for example, see \cite{7501477}). CNN gained popularity starting with AlexNet \cite{AlexNet} in the area of image processing, where they employed dropouts that increased the speed of CNN. For further information, please refer to the document on the ImageNet Large Scale Visual Recognition Challenge (ILSVRC) \cite{Russakovsky2015}. Further improvements have been made in the network architecture for improving the performance such as VGGNet \cite{VGG}, GoogLeNet \cite{GoogLeNet} etc. \cite{DLSurvey} gives an overview of the various deep learning algorithms and architectures available. \cite{DLAccess2} presents the various challenges that exist in conducting deep learning research while also briefing about the ongoing efforts and future trends of deep learning. 

For 3D graphical models, Wu et al. proposed ShapeNet  \cite{ShapeNet}, a dataset for volumetric shapes. As in the area of image processing, further improvements were made in the techniques for 3D graphical models either using a point-set representation (as in \cite{QiSMG17},\cite{Ravanbakhsh2017},\cite{PointNetA3},\cite{PointNet++}) or a Voxel-based representation (as in \cite{VoxNet},\cite{Qi2016VolumetricAM}).  While a few machine learning approaches perform well for images (sparse representations, manifold learning etc.), they are not popularly used for 3D data because such methods exploit the sturctured representation of data. For unstructured 3D data such as point sets etc., they may not perform very well.

\subsection{Summary}

Table \mbox{\ref{tab1}} provides a few details on the existing datasets for CAD models along with the number of models, the number of classes/class along with the category of classification. As can be seen from the table, most of them have only a very few number of models. Only ESB \mbox{\cite{ESB}}, which has 801 models, is available for public use, whereas the one in \mbox{\cite{Qin2014}} is not available for public use. From Table \mbox{\ref{tab1}}, it can also be observed that the datasets are either not sufficient enough or not available publicly for a deep-learning based approach.  Hence, there is a requirement to generate a dataset consisting of a few thousand models with labelled classification that can then be used for deep learning purpose.

In general, in the field of CAD/Engineering, very few problems have employed deep learning approach. Balu et al. \mbox{\cite{BaluLYKS16}} developed a voxel-based 3D CNN approach aimed at a framework for the design for manufacturability (DFM). Recently, Zhang et al. \mbox{\cite{ZHANG201812}} proposed FeatureNet, another voxel-based 3D CNN approach to learn machining or manufacturing features. 
For the task of classification of CAD models, a first of its kind deep learning approach was described in \mbox{\cite{Qin2014}} based on a proprietary dataset (7464 models from 28 categories). Using light field descriptor (LFD) for generating 2D images, they were converted to 1D feature vector using Zernike moments descriptor. This input vector was then fed into the deep neural network (DNN), and classification results were obtained. 

It can be clearly seen that the number of works are quite limited in the area of classification of CAD models. Even considering recent times, i.e., the last few yeas,  not much literature is available, to the best of our knowledge. Our aim is to bridge this gap by generating a dataset as well as using even further advances made in deep learning - CNN. 
Hence, in this paper, CNN, is employed for the functional classification of CAD models, perhaps for the first time. The focus is on the dataset `CADNET', which combines ESB, NDR as well as newly created 3D models. 

\begin{figure*}[t]
    \centering
    \includegraphics[height=8.5cm,width=17cm]{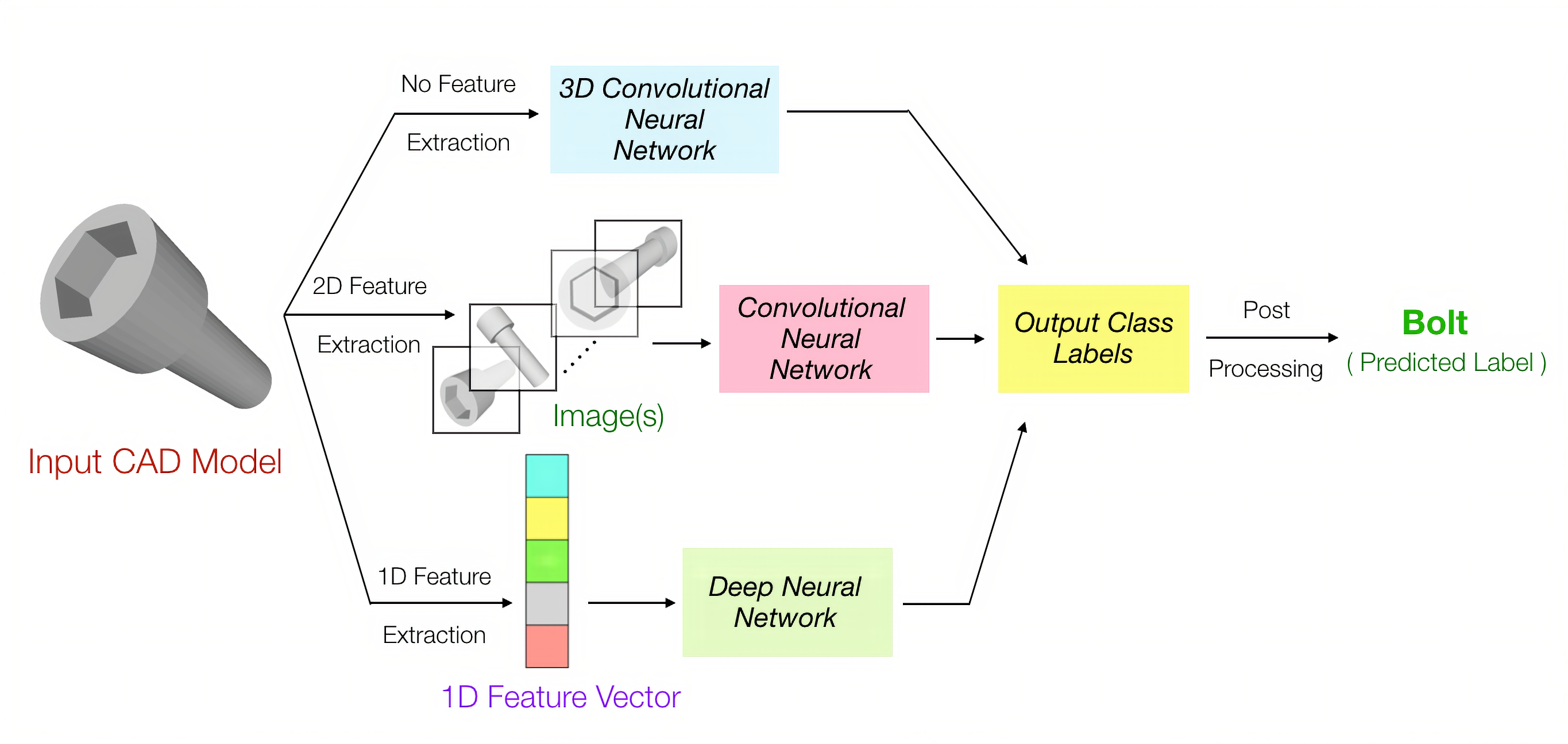}
    \caption{Illustrating the possible classification pipelines for 3D CAD Models using deep networks. We use the approach proposed in the `middle'.}
    \label{fig2}
\end{figure*} 

\section{Classification Pipeline}
\label{sec3}
The overall pipeline for classifying a 3D CAD model using deep learning can be broadly described as follows:
\begin{enumerate}
    \item Obtaining a dataset of 3D CAD Models which is suitable for training and testing a deep learning architecture. 
    \item Extracting features from a CAD model using a feature extraction method.
    \item Building a deep learning architecture that can efficiently be trained using the extracted representation as input.
    \item Post-processing of results (if any).
    \item Testing the network for performance.
\end{enumerate}
An overview of the possible classification pipelines is shown in Fig. \ref{fig2}. In the following sections, each step of the pipeline (Figure \ref{fig2}) is explained in greater detail.

\section{Dataset Preparation}
\label{sec4}

\subsection{Existing datasets}
A summary of existing datasets is shown in Table \ref{tab1}. As can be seen, some of them are based on the manufacturing process and not functionality. In the datasets that contain functional classification, most of them are not prepared for deep learning usage and hence contain very few models. The dataset from \cite{Qin2014}, however, is set up with the exclusive purpose of deep learning usage and hence appears to be very useful. It has 7464 models with 28 categories. The main limitation here is that it is a custom dataset and hence is not available for public usage. Also, the dataset has a high class imbalance, as observed from their paper. More recently, the ABC Dataset \cite{ABC} containing 1 million CAD objects has been made available. While it may sound promising, it is a mere collection of CAD objects which lacks any class or category information, thus making it unsuitable for a deep learning-based classification approach.

\renewcommand{\arraystretch}{1.5}
\begin{table*}[!t]
\resizebox{\textwidth}{!}{%
    \centering
    \begin{tabular}{|c|l|c||c|l|c||c|l|c|}
        \hline \textbf{Index} & \textbf{Category Name} & \textbf{\# of models} & \textbf{Index} & \textbf{Category Name} & \textbf{\# of models}& \textbf{Index} & \textbf{Category Name} & \textbf{\# of models} \\
        \hline 0 & 90\_degree\_elbows &  100 &  14 & Gear\_like\_Parts & 97 & 28 & Pulley\_Like\_Parts & 61 \\
         1 & BackDoors & 57 &   15 & Handles & 119 & 29 & Rectangular\_Housings & 70 \\
         2 & Bearing\_Blocks & 50  &   16 & Intersecting\_Pipes & 50 & 30 & Rocker\_Arms & 60 \\
         3 & Bearing\_Like\_Parts & 50  &  17 & L\_Blocks & 107 &  31 & Round\_Change\_At\_End & 51 \\
         4 & Bolt\_Like\_Parts & 111 &   18 & Long\_Machine\_Elements & 77 & 32 & Screws & 111 \\
         5 & Bracket\_like\_Parts & 27 &  19 & Long\_Pins & 104 & 33 & Simple\_Pipes & 66 \\
         6 & Clips & 54 &  20 & Machined\_Blocks & 59 & 34 & Slender\_Links & 60 \\
         7 & Contact\_Switches & 60  &   21 & Machined\_Plates & 99 & 35 & Slender\_Thin\_Plates & 62 \\
         8 & Container\_Like\_Parts & 60  &   22 & Motor\_Bodies & 58 & 36 & Small\_Machined\_Blocks & 62 \\
         9 & Contoured\_Surfaces & 55  &  23 & Non-90\_degree\_elbows & 108 & 37 & Spoked\_Wheels & 57 \\
         10 & Curved\_Housings & 51  &  24 & Nuts & 125  & 38 & Springs & 55 \\
         11 & Cylindrical\_Parts & 94  & 25 & Oil\_Pans & 58 & 39 & Thick\_Plates & 82 \\
         12 & Discs & 163  & 26 & Posts & 109 & 40 & Thin\_Plates & 83 \\
         13 & Flange\_Like\_Parts & 109  & 27 & Prismatic\_Stock & 86 &  41 & T-shaped\_parts & 65 \\
         & & & & & &  42 & U-shaped\_parts & 75 \\  \hline
           &&&&&& & \textbf{Total \# of models} & \textbf{3317} \\
        \hline
    \end{tabular}
    }
    \caption{\label{tab2} Details of the developed `CADNET' dataset - Name of the category and the number of models in each of them.}
\end{table*}
\renewcommand{\arraystretch}{1} 

\subsection{`CADNET' - A collection of 3D CAD models}
In the NDR dataset (based on functional classification) \cite{NDR}, there are 70 models over 10 classes. Although insufficient for training a deep neural network in itself, it is publicly available and the classification is based on the functionality. In the ESB dataset \cite{ESB}, there are 801 models over 42 categories (excluding the objects from `Miscellaneous' class), also classified based on their functionality. ESB is a reasonably sized dataset, contains objects from many categories and is publicly available. The models from these two datasets are collected and have been combined into a single dataset - after checking for duplicates, overlapping classes etc. This resulted in a collection of 868 3D CAD models over 43 categories. Although we obtain a decent average of 20 models per-class, this collection is quite imbalanced with the number of models per category ranging from as low as 4 to the highest being 61. Herein arises a need to generate more data in order to increase the size of the dataset as well as to cover the imbalance as much as possible.

The procedure adopted in order to achieve this is as follows. By observing the 3D objects in each category, an overview of the 3D designs is obtained. Using this knowledge, a few representative models are parametrically designed in Autodesk Fusion360 software - for each class. Following this, more 3D models are generated via a python script linked to the Autodesk Fusion360 API, for various sets of parameter values. 

For example, in order to model a cuboid parametrically, one needs three parameters - for the three dimensions (say $l$, $b$, $h$). For different sets of values for $l$, $b$ \& $h$, we get corresponding   cuboids. This process essentially generates multiple variants of a certain category by using many sets of parameter values. In this way, many training examples are created. Every generated 3D model is then verified for correctness. Repeating this process for every category in the collection mentioned above, a dataset of 3317 3D CAD objects over 43 categories is obtained. We refer to this dataset as `CADNET'. The dataset is made available at \url{https://github.com/bharadwaj-manda/CADNET_Dataset} 

\begin{figure}[h]
    \centering
    \includegraphics[width=8cm,height=7cm]{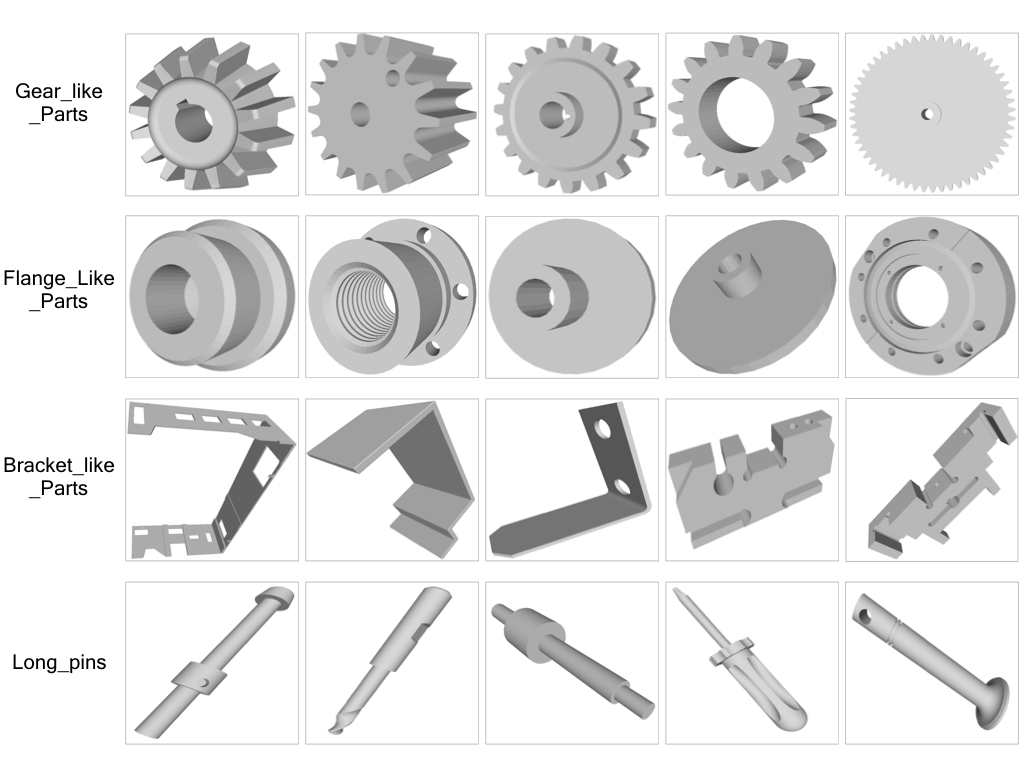}
    \caption{Sample models from the `CADNET' dataset}
    \label{fig7}
\end{figure}

Table \ref{tab2} shows the details of CADNET, which gives the category name and number of models in each of them. A few sample models from CADNET are shown in Figure \ref{fig7}. The dataset CADNET now has a significantly large number of models, which can be used in a deep learning setting. The classes are still not balanced, and the method adopted to tackle this is presented in section \ref{sec_classimb}.

\section{Feature extraction and CNN-based classification}

In the recent past, the emergence of other deep learning techniques has revolutionized the use of machine learning for various research domains. A significant breakthrough in the field of image classification has been achieved in \cite{AlexNet}, where Convolutional Neural Networks (CNNs) are used for the task of classifying more than 1 million images belonging to 1000 classes, as a part of the ImageNet Large Scale Visual Recognition Challenge (ILSVRC) in the year 2012.  This motivated us to employ CNN for CADNET.

\subsection{Feature Extraction - Light field descriptor (LFD) with weighted views}
\label{sec5}
 
CNN has been demonstrated extensively on 2D images as inputs. As our inputs are 3D models, it is not very evident on how to convert them to images. Nevertheless, there exists a popular approach called light field descriptor (LFD) that generates a set of images from a 3D model that can then be used for  2D CNN. The idea of LFD, as described by \cite{LFD}, uses 20 cameras placed at the vertices of a regular dodecahedron to capture images of the 3D model from various views. We use this method to obtain 20 images for every CAD model from our dataset. Each of these images is then assigned the same class label as that of the corresponding CAD model. This process is repeated for every model in the dataset. Now, we use this set of annotated images as the input data for the convolutional neural network.

It should be noted here, however, that this LFD technique is not the same as the one used in \cite{Qin2014}. In \cite{Qin2014}, 10 light fields are created for each 3D model, and 10 images are extracted from each light field. Thus, in effect, 100 images are extracted for a single 3D Model which are further processed by a Zernike Moments descriptor that produces a 1D feature vector. In our case, we greatly simplify the process by using just 20 images per 3D model (i.e. one light field). The reasons for using only 20 images over 100 images as mentioned in \mbox{\cite{Qin2014}} are because \\
(1) Chen et. al \mbox{\cite{LFD}} showed that 20 images are sufficient and using more than that leads to  redundancy.  We also observed a similar trend when more number of images were used, and,  \\
(2) Using 20 images also takes much lesser time for computation. \\ 
On top of that, we employ a post-processing scheme (as will be discussed in Section \ref{subsec:post}), a machine learning algorithm, which learns the influence of each viewing direction on the output prediction. Thus, in effect, we use a weighted LFD approach to extract images from the 3D model, while also knowing the effect of the individual viewing direction.

\begin{figure}[t]
    \centering
    \includegraphics[height=3.5cm,width=7.5cm]{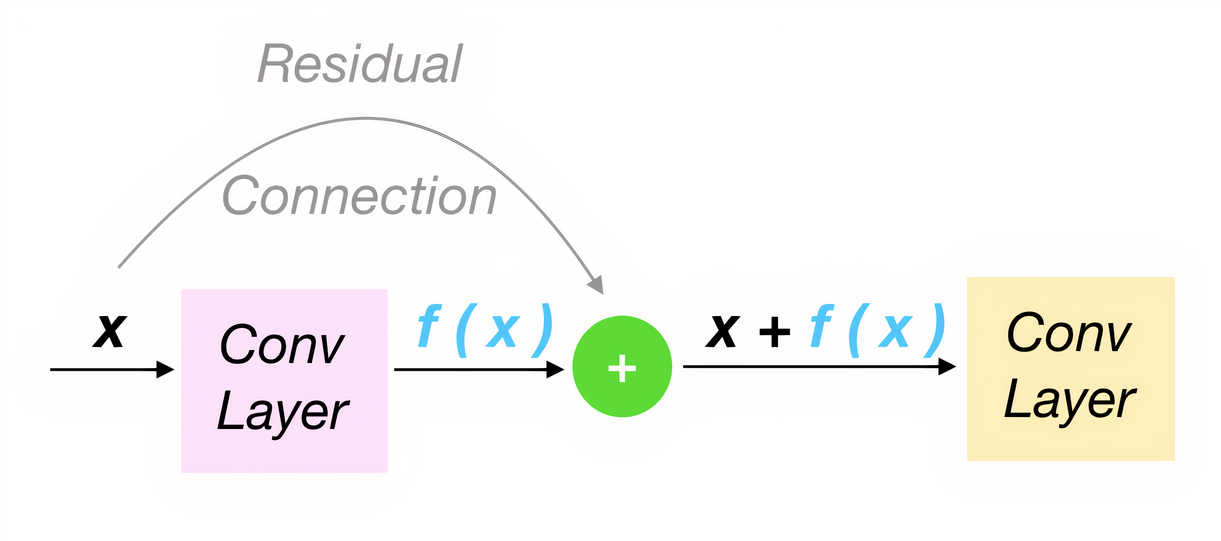}
    \caption{Illustrating the idea of residual connections \cite{ResNet}}
    \label{fig4}
\end{figure}

\section{Network Architecture}
\label{sec6}
\begin{figure}[t]
    \centering
    \includegraphics[width=3.5cm,height=13cm]{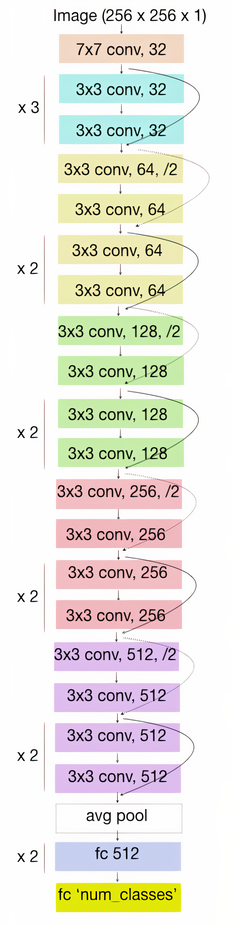}
    \caption{Proposed CNN Architecture. This network requires lesser number of filters thereby reducing the number of parameters. Details are presented in Section VI. }
    \label{fig3}
\end{figure}

As part of the ILSVRC challenge, every year newer, deeper, and more efficient architectures have been proposed, outperforming the previous ones. AlexNet \cite{AlexNet} was the first architecture to have shown tremendous improvement in classification performance using CNNs when compared to earlier methods. VGGNet \cite{VGG} has bettered the classification results as compared to AlexNet. Considered to be very deep at the time of its introduction, VGGNet takes an enormous amount of time to train. GoogLeNet \cite{GoogLeNet} (sometimes referred to as InceptionNet v1) provided further improvement in accuracy by using filters with multiple sizes operating on the same level, i.e. using a `wider' network rather than a `deeper' one. The winner of the 2015 edition of the challenge is an architecture proposed by \cite{ResNet}, popularly known as ResNet. It uses the idea of residual connections, that helps in faster training of deeper networks. Inspired by this idea of residual connections (see Fig. \ref{fig4}), and the advantages it offers for faster and effective training of deep networks, we build a 35-layer CNN (see Fig. \ref{fig3}). As discussed in Section \ref{sec5}, 20 images are extracted from each 3D CAD model in the dataset (both for train set and test set). For each of these 20 images, the class label of the corresponding input model is assigned to them. The images from the training set are then used for training the CNN.

\subsection{Input layer}
The input to the network is an image of size 256*256*1. Our input convolutional layer consists of 32 filters, each of which performs a 7*7 convolution operation on the input image. The output of the convolutions is then passed through a series of hidden convolutional layers before reaching the final output layer. The activation function used is ReLU.

\subsection{Hidden layers and Residual connections}
\label{sec6.2}
The hidden layers consist of many residual blocks. In general, when we have a series of convolutional layers, the input of each layer is simply the output of the previous layer. However, when we have a residual connection between two layers, the input of each layer will be a summation of the output of the previous layer and the value from the residual connection (See Figure \ref{fig4} for illustration). The hidden layers are organized as shown in Figure \ref{fig3}. The arrangement of the hidden layers is explained as follows.

There are 5 `groups' of hidden layers (each indicated by a different colour in Figure \ref{fig3}). Each group has three residual blocks. Residual connections exist between consecutive residual blocks. Each residual block has two hidden layers, and hence there are six hidden layers per group. Batch normalization is applied at the beginning of every residual block. Solid lines indicate that the number of filters remains the same, while dashed lines indicate that there is an increase in the number of filters by a factor of 2. The number of filters in each layer and the filter size are mentioned across each layer in the Figure. Convolutions are performed using stride = 1. Some hidden layers are indicated by a `/2' in the Figure. In such layers, two additional operations are performed ahead of the batch normalization - (1) max-pooling of size 2*2, (2) 1*1 convolution operation with stride = 2. ReLU activations are used in all hidden layers. 

Batch normalization is applied to the output of the last layer of the last group. It is then fed into a pooling layer that performs average pooling of size 4*4. The output is then flattened into a 1D vector and is then fed into a series of two fully connected layers, each of 512 nodes. ReLU activation is used in both these layers. Dropouts with a probability value of 0.25 are applied in these two layers in order to enhance the prediction accuracy and to avoid any overfitting.

The total number of hidden layers = 5*6 (convolutional layers) + 1 average pooling layer + 2 fully connected layers = 33 layers.

It is to be noted here, that although the idea of residual connections is adopted from \cite{ResNet}, the proposed network architecture differs significantly from that of ResNet. These differences arise primarily due to the differences in the nature of the data involved. ResNet, built for the purpose of ImageNet Classification, deals with images that contain real-world graphical objects and each datum consists of many details. In our case, we specifically focus on the images that are extracted from Engineering/CAD Models, which have relatively lesser information as compared to the images from ImageNet dataset. The images are also single-channel, unlike ImageNet, where each image is an RGB (3-channel) image. It should also be noted that although the images extracted from LFD are of size 256*256, it only serves as an outer boundary for the 3D objects. The actual image is contained within this square, and a significant portion of the image is empty, unlike the images from ImageNet. Hence, 
\begin{enumerate}
	\item The proposed network requires a fewer number of filters to capture the features in the initial layer (32) as opposed to ResNet (64).
	\item We require a lesser number of filters (in the hidden layers) compared to ResNet to extract the features from the images. Hence we have six layers of 32, 64, 128, 256 and 512 (= 5952 filters) as opposed to 6 layers of 64, 8 layers of 128, 10 layers of 256 and 6 layers of 512 (= 7040 filters) in ResNet. 
	\item This, in turn, reduces the number of parameters (by about a million in the network)
\end{enumerate}

\begin{figure}[t]
    \centering
    \includegraphics[height=7cm,width=9cm]{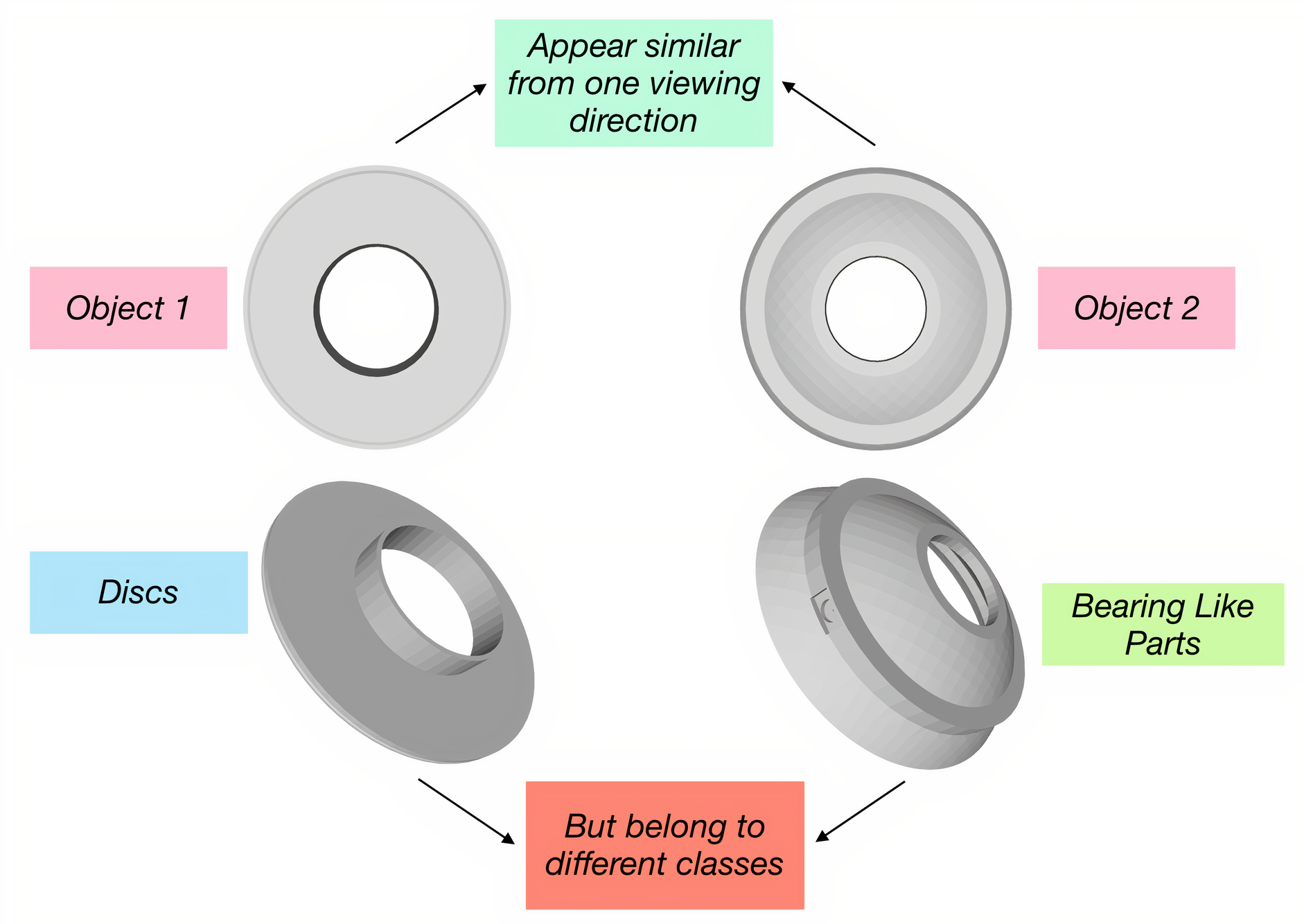}
    \caption{Need for a post processing scheme to reduce error due to misclassifications}
    \label{post}
\end{figure} 

\subsection{Output Layer}
The output from the last hidden layer is then fed into a fully connected layer, with the number of nodes equal to the number of classes. In our case, it is 43 for CADNET. The activation function used here is softmax. The output of this layer is a 1D vector of size 43. The values of this vector indicate probability values. Based on the highest probability value, the class label for the input image is obtained.

\begin{figure*}[!t]
    \centering
    \includegraphics[height=5cm,width=18cm]{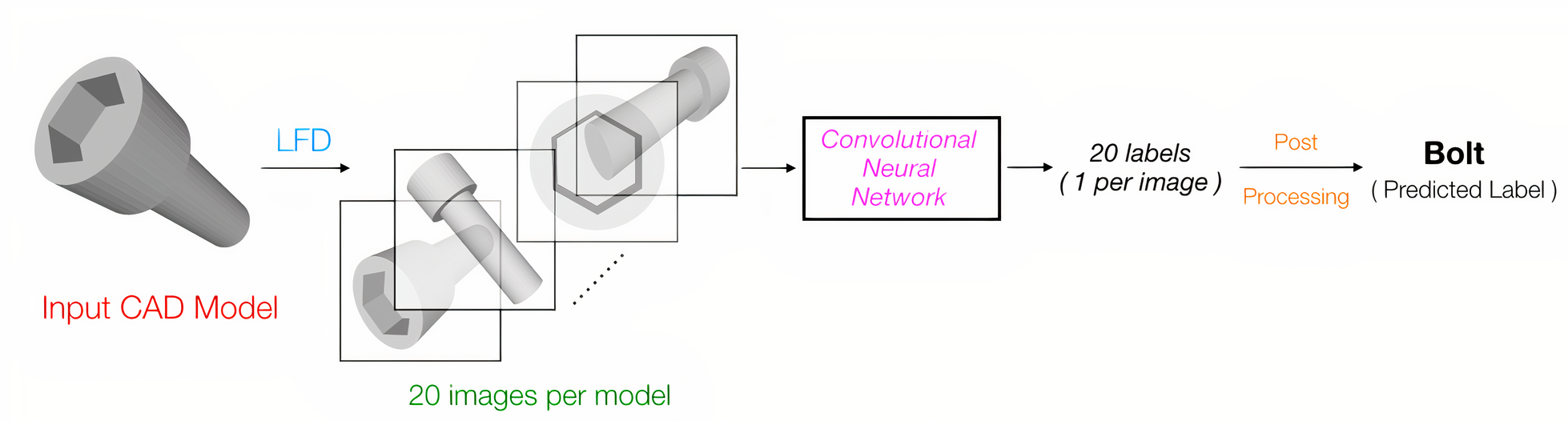}
    \caption{Pipeline of our approach for the classification of a given CAD model using LFD, convolution and post-processing (weighting using boosting approaches).}
    \label{pipeline}
\end{figure*} 

\subsection{Post Processing}
\label{subsec:post}
The output of the network is a vector of 43 probability values for every input image. Generally, the class with the maximum probability value is chosen to be the class label. In this case, however, the machine learning algorithms such as XGBoost \cite{XGBoost}, and CatBoost \cite{CatBoost} are used to do this. The reason is that LFD is a view-based method, and some images are misclassified because models from different classes appear similar from a certain viewing direction (Refer to Figure \ref{post}). In order to reduce such misclassifications and thus enhance the prediction accuracy, we process the probability values using XGBoost and CatBoost. These algorithms output a single class label for the image by learning the effect of different viewing directions. Now, since each 3D CAD model has 20 images, we would have 20 labels per model. A majority vote of these 20 values is then taken, and a single label per model is obtained.

\section{Implementation Details}
\label{sec7}

The overall pipeline of our implementation for a CAD model classification is depicted in Figure \ref{pipeline} with LFD as the feature extraction, with a CNN using the network architecture as described in Section \ref{sec6}, and with post-processing as discussed in Section \ref{subsec:post}.

\subsection{Training and test set}\label{sec_trainset}
In case of large datasets such as \cite{Russakovsky2015} for images, \cite{ShapeNet} for 3D shapes etc., the process of splitting the data into training and test sets is quite straight forward - a percentage of samples are chosen randomly for the test set and the remaining for the training set. This idea generally works because the number of per-class models available for training is high. Also, all the intra-class variations are more or less sure to be captured, owing to the size of the dataset. The same idea is applied for CADNET, with a train-test split of 80-20, which results in 2654 models for training and 663 models for testing. A lower split \% for training set resulted in faster training, but at the cost of lesser accuracy - for want of more training data. Higher split \% for training set resulted in overfitting the training data, while also taking much longer to train. The 80-20 split used in our training methodology is not arbitrary. It is as per the standard Pareto Principle, which is quite widely used in literature, and our experiments seemed only to reconfirm this. Hence, we presented the results of the 80-20 split, which yields the best classification accuracy.


\begin{figure*}[!h]
    \centering
    
    \begin{subfigure}[b]{0.45\textwidth}
        \centering
        \includegraphics[height=5cm,width=1\textwidth]{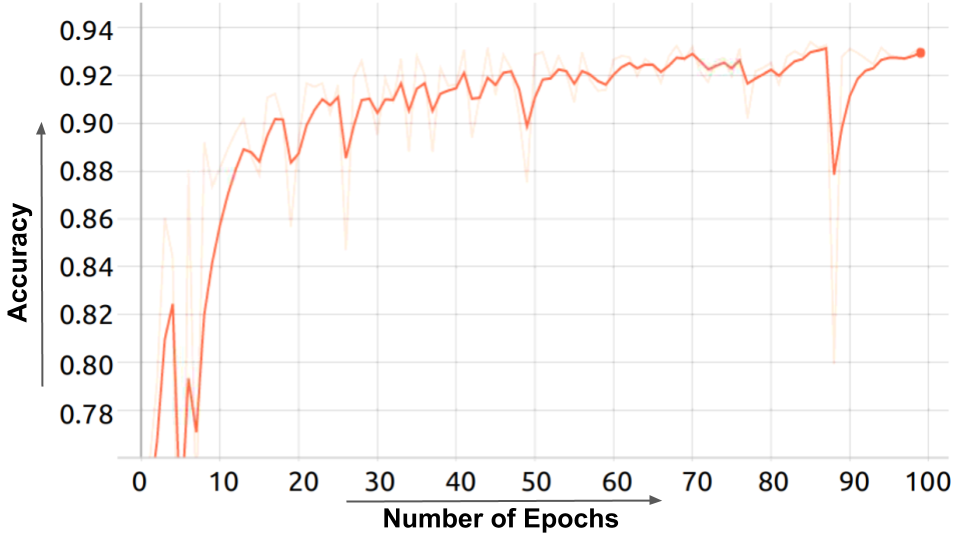}
        \caption{}
        \label{fig6a}
    \end{subfigure}
    \begin{subfigure}[b]{0.45\textwidth}
        \centering
        \includegraphics[height=5cm,width=1\textwidth]{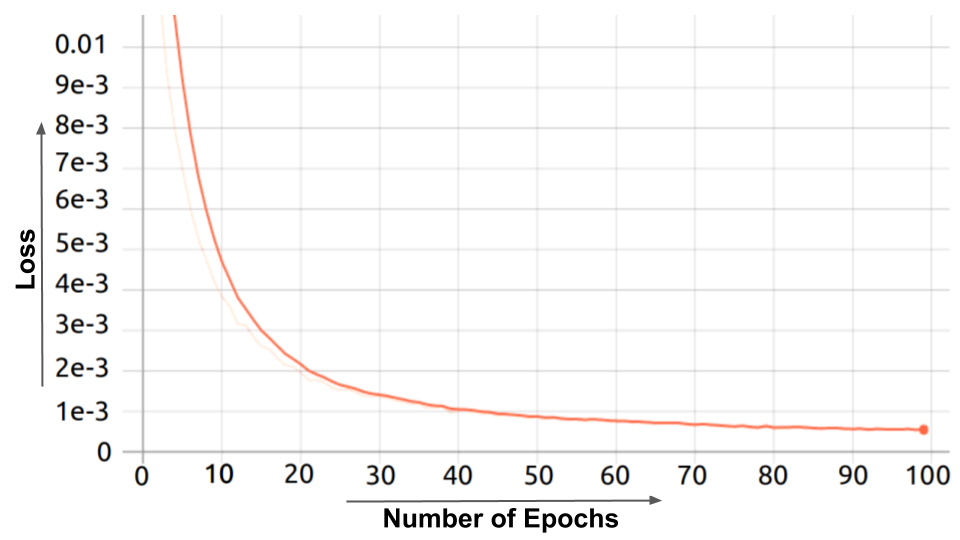}
        \caption{}
        \label{fig6b}
    \end{subfigure}
    \centering
    \caption{Plots of Accuracy and Test Loss over the training period (Number of epochs).  (a) CNN accuracy increases as training time progresses; (b) CNN test loss decreases as training time progresses}
\end{figure*}

\subsection{Addressing class imbalance}\label{sec_classimb}

The problem of class imbalance still exists in this dataset (see Table \ref{tab2} for details). We tackle this issue by using the idea from \cite{LogReg}, during the prediction phase of the neural network. We estimate class weights for the unbalanced dataset. The weight of each class is simply the ratio of the number of samples to the product of classes and the bin-count of the class labels.

\subsection{Hyperparameter Tuning, Loss function \& Optimization}
Training a neural network is a tedious task because of the many decisions involved such as choice of performance metrics, hyper-parameters, loss function, debugging strategies etc. Our choices are mainly based on heuristics (\cite{Goodfellow}, \cite{Bengio2012PracticalRF}, \cite{Larochelle2009ExploringSF} ) and are backed by experimental verification.  

After various experiments based on heuristics, we choose a learning rate of 0.001 for training our network. The back-propagation algorithm \cite{Rumelhart1986LearningRB} is used for training the neural network. There are various numerical optimization algorithms available (\cite{Ruder2016AnOO}). We adopt a mini-batch training scheme that uses more than one training example but less than the total number of examples at once. The training examples are split into many batches with 20 examples per batch.

Since the task at hand is multi-class classification, we use the categorical cross-entropy loss function, and the Adam optimization algorithm \cite{Adam} is used to minimize this loss function. The CNN is trained for 100 epochs. For regularization, there are various methods in practice (\cite{Goodfellow3}) such as enforcing norm penalties, early stopping, etc. In our case, we use the idea of dropouts \cite{Dropout} with a probability value of 0.25 in for the fully connected layers.

\subsection{Coding Framework and System Configuration}
For implementing our neural network, we use Python3 with Keras \cite{Keras} and Tensorflow \cite{TFPaper} (gpu-version 1.11.0 \cite{TF}). In order to implement the XGBoost and CatBoost algorithms, we use Python3, and sklearn \cite{sklearn}. 

All the implementations are carried out on a system running Ubuntu 18.04 Operating System. The system has an Intel Core i7-4930K CPU with 64GB RAM and an NVIDIA GeForce GTX 1080Ti GPU with 11GB RAM.

\renewcommand{\arraystretch}{1.3}
\begin{table*}[!t]
\resizebox{\textwidth}{!}{%
    \centering
    \begin{tabular}{|c|c|P{3cm}|P{3cm}|c|}
        \hline \textbf{Index} & \textbf{Category} & \textbf{\# of models in test set}&\textbf{\# of misclassified models} & \textbf{Misclassified as}\\
        \hline 
  		0 & 90\_degree\_elbows & 20 & 1 & Simple\_Pipes \\ \hline
  		4 &	Bolt\_Like\_Parts & 22 & 3 & Posts, Screws \\ \hline
  		5 & Bracket\_like\_Parts & 5 & 1 & Small\_Machined\_Blocks \\ \hline
  		10 & Curved\_Housings & 10 & 2 & 90\_degree\_elbows \\ \hline
  		11&	Cylindrical\_Parts	&18	&1	&Posts \\ \hline
  		12&	Discs	&32	&2	&Flange\_Like\_Parts \\ \hline
  		20&	Machined\_Blocks	&11	&1	&Rectangular\_Housings \\ \hline
  		21&	Machined\_Plates	&19&	2&	Thick\_Plates, Thin\_Plates \\ \hline
  		24&	Nuts&	25	&1&	Discs \\ \hline
  		29&	Rectangular\_Housings &	14	&2	&Thin\_Plates, Long\_Machine\_Elements \\ \hline 
  		32&	Screws	&22	&2	&Bolt\_Like\_Parts \\ \hline
  		34&	Slender\_Links	&12	&3	&Long\_Machine\_Elements, Thin\_Plates \\ \hline
  		36&	Small\_Machined\_Blocks&	12	&1	&Long\_Machine\_Elements \\ \hline
  		37&	Spoked\_Wheels	&11	&1	&Gear\_like\_Parts \\ \hline
  		40&	Thick\_Plates	&16	&2	&Rectangular\_Housings, Thin\_Plates \\ \hline
  		41&	Thin\_Plates &	13	&4	&Rectangular\_Housings, Machined\_Plates \\ 	\hline
  		 &	Remaining classes &	401	&0	& -- \\ 	\hline
        \hline & \textbf{Total} & \textbf{663} & \textbf{29} & \\
        \hline
    \end{tabular}
    }
    \caption{\label{tab4} Misclassification Results on CADNET}
\end{table*}
\renewcommand{\arraystretch}{1}

\section{Results and Discussion}
\label{sec:res}
Our CNN classifier is evaluated for performance on the CADNET dataset, and the results are reported in this section. Training and test sets are chosen, as discussed in Section \ref{sec_trainset}. As there exists class imbalance, it is addressed, as discussed in Section \ref{sec_classimb}. The obtained accuracy is then put into perspective. 
Our results are also compared with various other features and their accuracies obtained using a deep neural network. 


\subsection{Results on CADNET dataset}
The CNN training time is approximately 30h, due to a large number of inputs (2654*20 images) that the CNN has to process. The accuracy computed at the output layer of the CNN is 93.41\%. The accuracy is further improved using XGBoost/CatBoost algorithm as in Section \ref{subsec:post}. Roughly, the time taken for XGBoost is 15s and for CatBoost is 12s. With XGBoost, the obtained accuracy is 95.63\% with 29 models misclassified out of 663 models in the test set. Similar results are obtained using CatBoost - 95.47\% with 30 misclassifications. Figures \ref{fig6a} and \ref{fig6b} show the plots of CNN accuracy and test loss respectively with respect to the number of epochs when trained on CADNET. We report the results using XGBoost since it obtains a higher numeric value for accuracy as compared to CatBoost.

\begin{figure}[t]
    \centering
    \includegraphics[height=5cm,width=8cm]{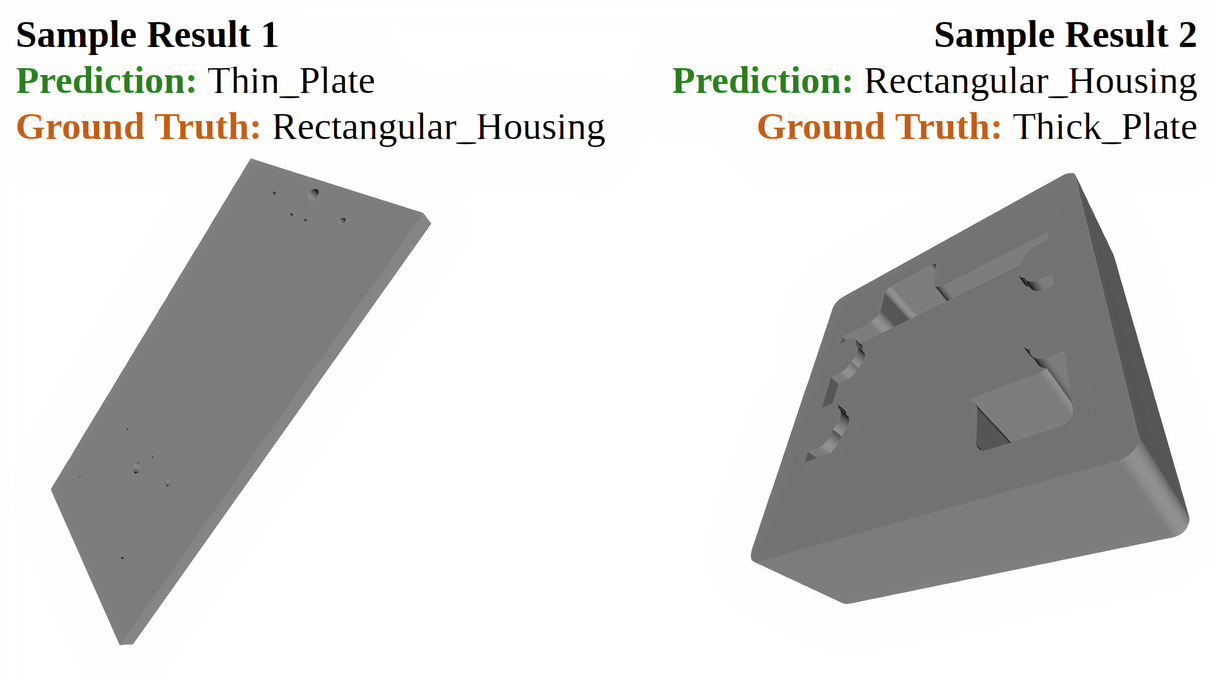}
    \caption{Sample misclassification results of the proposed method when trained on CADNET.}
    \label{fig6}
\end{figure}

\begin{figure*}[t]
	   \centering
	   \includegraphics[scale=1.4]{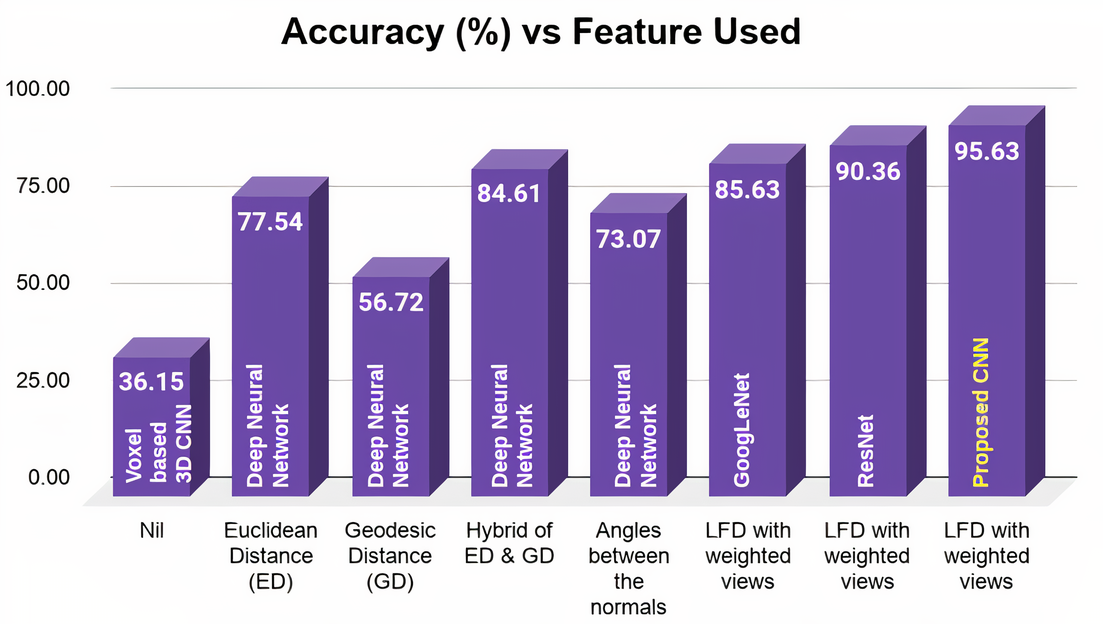}
       \caption{Comparing various feature-classifier combinations on CADNET}
       \label{fig10}
\end{figure*}

Table \ref{tab4} shows per-class classification results after applying boosting. There are a total of 663 models in the test set, across 43 categories. Out of these, 27 classes are perfectly classified without a single misclassification (401 models); for the majority of the remaining classes, there are just 1 or 2 models that are misclassified. For a majority of these models, the misclassification results are quite understandable since the class predicted by the proposed architecture and the actual class have a lot of similarities. For example, a model from the class Thick\_Plates is wrongly identified as Thin\_Plates, since both classes are Plates. Also, 2 models from Bolt\_Like\_Parts are classified as Screws, and 2 models from Screws are identified as Bolts. A Spoked\_Wheel is identified as a Gear and so forth. These classes are pretty similar, and the resulting misclassifications are understandable. This is due to LFD being a view-based technique. For these models, when the extracted images by LFD are visualized, they look very much similar to each other and hence the wrong predictions. 

Regarding the apparently non-obvious misclassifications, when the models are visualized, they look very different from the other objects of the same class. In fact, they look similar to some objects of the predicted class. For instance, Figure \ref{fig6} shows some visual results of the wrongly classified models. The first object is from the class "Rectangular\_Housing" which it is misclassified as a "Thin\_Plates". Similarly, for the second object, the prediction is "Rectangular\_Housing", while the object is a "Thick\_Plate". It is easy to see why these misclassifications occur, as the models look very similar to some objects of the predicted class.

\subsection{Comparison with other methods on CADNET using deep networks}
There is no information available to directly compare the proposed CNN-based approach on CADNET.  As there was no deep network-based approach employed on CADNET, we also implemented a deep neural network (DNN). We need to select and extract features from each input 3D CAD model such that the extracted representation can capture the essential information from the input model. As an initial trial, a naive Voxel-based 3D CNN approach is used. It is a 14 layer network which consists of a series of 3D Convolution and 3D MaxPool layers followed by two dense layers. This naive approach of directly using a Voxel-based 3D CNN, performed very poorly (classification accuracy is 36\%), which is to be expected due to the presence of features such as holes etc. in 3D CAD models. Also, the presence of many empty/sparse voxels arising due to sparse nature of the point-sets of CAD models do not help in obtaining a better accuracy.

Further experiments are carried out using an 8-layer, fully connected, deep neural network (DNN). Extensive 3D feature extraction techniques have been proposed in \cite{ShapeSearch}, \cite{ShapeSearch2}, \cite{ShapeSearch3} and \cite{ShapeSearch4}. For our work, we have tried various geometry-based feature extraction methods such as Euclidean distance between points, geodesic distance, a hybrid of Euclidean and geodesic distances, the angle between normals etc. Using these methods, we obtained a 1D vector representation of the 3D shape and then fed it as an input vector to the deep neural network. 

Figure \ref{fig10} indicates the accuracy results using various feature-classifier combinations on CADNET. It can be observed that the best accuracy obtained is only 84.61\% using 3D signatures, while the naive Voxel-based 3D CNN performed very poorly. On the other hand, our method of using LFD  with CNN, along with combining weighted views yielded much higher accuracy. The proposed network architecture is also compared with state-of-the-art CNN architectures such as GoogLeNet \cite{GoogLeNet} and ResNet \cite{ResNet}. The proposed network yielded a maximum accuracy of 95.63\%, much higher than the next best one, while also taking much lesser time to train.

\begin{figure}[t]
	   \centering
       \includegraphics[scale=1.2]{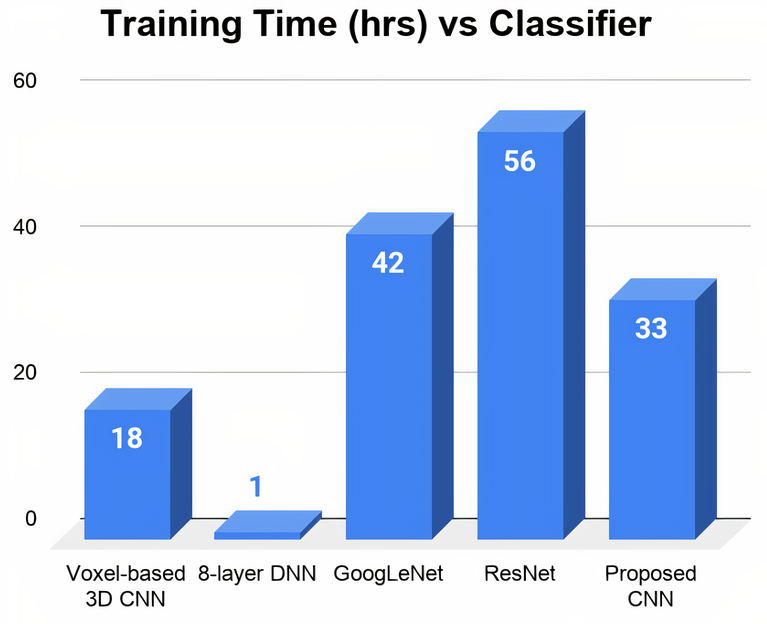}
       \caption{Comparing the training time of various networks when trained on CADNET}
       \label{fig11}
\end{figure}

%

\subsection{Comparison with deep learning approaches used for Graphical models}

As mentioned in Section \ref{sec2}, most deep learning approaches for graphical models either use a point-set representation or a Voxel-based representation. It may be noted that the point-set representation of CAD models is sparse and non-uniform, thus leading to many empty voxels in the voxel-based representation. Hence, a direct 3D CNN on a voxel-based representation may not work well for the classification task of an entire 3D CAD model (substantiated with results from Figure \ref{fig10}). From this, it can be seen that such methodologies adopted for 3D graphical models need not perform very well on 3D Engineering/CAD models. A few approaches, however, such as the Multi-View CNN (MVCNN)\cite{MVCNN}, use a view-based method, i.e. use images of the 3D models. Since such an approach appears to work well on CAD models, we experimented with this approach on CADNET. 

MVCNN uses two camera setups - (1) 12 views (one view for every 30$^0$ from 0$^0$ to 360$^0$), under the assumption that the input shapes are upright oriented along a consistent axis and (2) 80 views: 20 views that are obtained from viewpoints at the 20 vertices of an icosahedron enclosing the shape and then from each viewpoint, using 0$^0$, 90$^0$, 180$^0$ and 270$^0$ rotation along the axis passing through the viewpoint and the object centroid. No prior assumption is made regarding the orientation of the object in this case. A TensorFlow implementation of MVCNN with the first camera setup is available publicly on GitHub \cite{mvcnn_github}. The method uses 12 images, extracted from 12 viewing directions, and these images are fed into the network which is trained on CADNET for of 100 epochs (default according to the paper). 

The accuracy obtained using this method on CADNET is only 58.75\% after training for 100 epochs. By further increasing the number of epochs, a maximum accuracy of 61.25\% is obtained at around 170 epochs, after which no further improvement is noticed. In fact, after 200 epochs, the model begins to overfit the training data, and the test accuracy begins to reduce. This is to be expected due to several reasons. Firstly, MVCNN with the first camera setup uses 12 views, under the assumption that the input shapes are upright oriented along a consistent axis. While this assumption might hold true for datasets such as ModelNet, the same cannot be said for CAD datasets where the objects consist of volumetric features that are not always oriented along a standard axis. Also, using just these 12 views might not be sufficient enough for training.
Secondly, the network architecture is not very deep - only 5 conv layers, 1 view pooling layer followed by 3 fully connected (fc) layers (very much like AlexNet which also, incidentally, has 5 conv and 3 fc layers). Considering the fact that deeper architectures for images such ResNet have shown significant improvement in results as compared to AlexNet, we have also experimented with a modified architecture for MVCNN that is \lq ResNet-like'. This resulted in an accuracy of  72.33\% - indicating a strong influence of the network architecture on the classification accuracy. 

\begin{figure}[!t]
    \centering
    \includegraphics[scale=.25]{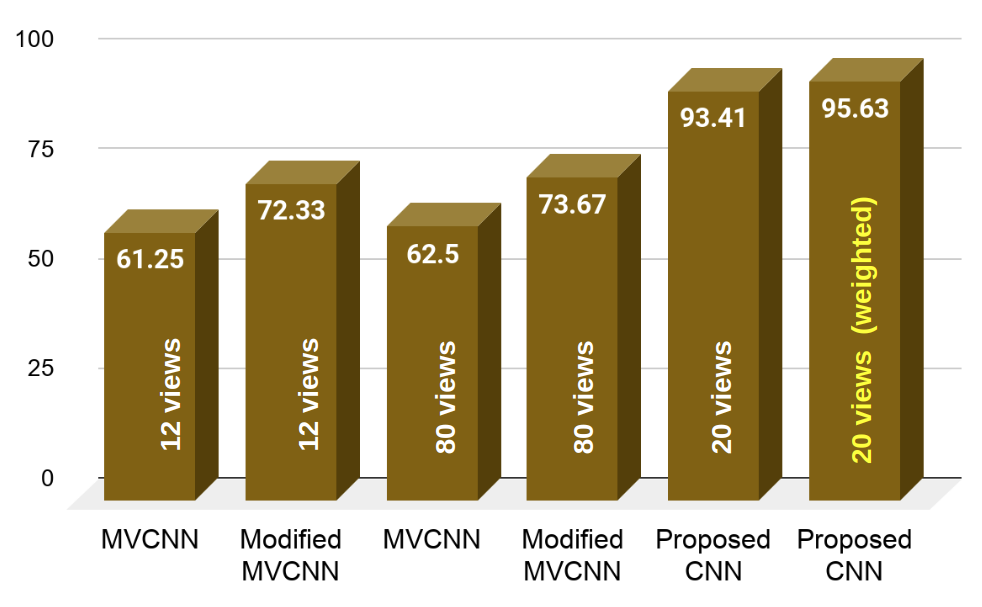}
    \caption{Comparing the accuracies of view-based deep learning techniques on CADNET}
    \label{fig12}
\end{figure}

On similar lines, extending the available TensorFlow implementation for the second camera setup, the MVCNN architecture is trained on CADNET once again for 100 epochs. The obtained test accuracy is only 41.25\% when trained for 100 epochs. By further training, an accuracy value of 62.5\% is obtained at around 260 epochs, beyond which the model begins to overfit the training data. The reason for the failure of MVCNN to perform well on CADNET, even using the second camera setup - without any assumption regarding the orientation of the objects, could be that ModelNet is a relatively well-balanced dataset, with many more number of training examples as compared to CADNET. Since class imbalance is not well-handled in this method, it fails to perform well on CADNET, in which class imbalance is present. Since our method takes into account the imbalanced nature of the dataset, it performs way better than MVCNN on CADNET. In addition to this, there is every possibility that using such a large number of views could lead to overfitting, as not much additional information is obtained through these images. This is evident in \cite{MVCNN} where the improvement in accuracy is less than 1\%. Like in the first case, we have also experimented with a modified \lq ResNet-like' architecture for MVCNN which resulted in an accuracy of 73.67\% - once again indicating the strong influence of the network architecture on the classification accuracy.

In summary (see Figure \ref{fig12}), our proposed network architecture, even with plain LFD (without weighted views), performs much better than the highest obtained accuracy using the techniques mentioned above on CADNET, resulting in an accuracy of 93.41\% (next highest is 73.67\%), despite using a ResNet-like architecture. The reason for this method to perform well is due to the modifications that have been done to the architecture, to suit CAD model images (as elaborated in Section \ref{sec6.2}). Furthermore, by using the proposed weighted LFD views scheme, the accuracy is improved further to 95.63\%. From these experiments, it can be concluded that the proposed network architecture, along with the view-weights and class-weights approach achieves the best performance on CADNET. 

\subsection{Limitations and Possible future work}
The scope of this work is limited to 3D CAD Mesh models. Other kinds of inputs, such as images or even 3D point sets etc., are not handled by the proposed approach. It is worth exploring to consider building a unified network architecture to process multiple input formats. Also, one possible way to improve the results could be to use a much deeper network and with many more filters in each layer.  

Also, when the dataset is made open, users could contribute towards enhancing the dataset, which in turn can be used to increase the performance of the network. As an extension of this work, an automatic CNN-based 3D CAD model retrieval system can be developed. The input queries to the search engine can be compared against the models from CADNET database, using the classification results from the proposed CNN. The current work could also be extended to a CAD assembly model retrieval \mbox{\cite{CAD_assem}} and sketch-based retrieval of CAD models \mbox{\cite{Qin2017}}.

\section{Conclusion}
\label{sec:concl}
We built a collection of 3D CAD (Engineering) models with functional classification, termed as CADNET,  using the available data from existing datasets ESB and NDR, and augmenting them with manually generated models. A convolutional neural network (CNN) classifier for 3D CAD models was then built, perhaps for the first time. It was observed that processing the 3D CAD models directly into a 3D CNN yielded poor results. Hence, light field descriptor (LFD) was then used for extracting features from a CAD model, and the obtained images were fed into the proposed CNN. A residual network architecture for CAD models with much lesser number of filters (thereby reducing the number of parameters and the time for training) was also proposed.  We also observed that 20 images per CAD model was sufficient. The problem of class imbalance was addressed by using a class-weight approach. Post-processing of the CNN results was done using XGBoost / CatBoost. It was also shown that proposed approach resulted in the highest classification accuracy when compared to other features/networks. Possibilities of extending this work to related research problems have also been discussed.

\section*{Acknowledgments}
Thanks are due to the teams of Purdue ESB and National Design Repository, for making their data publicly available. 

\footnotesize
\bibliographystyle{unsrt}
\bibliography{refs}
\end{document}